\documentclass[letterpaper, 10 pt, journal, twoside]{ieeetran}
\pdfoutput=1
\IEEEoverridecommandlockouts                              

\usepackage{graphics} 
\usepackage{graphicx}
\usepackage{subfigure}
\usepackage{mathtools}
\usepackage{amsmath}
\usepackage{amssymb}
\usepackage{amsthm}
\usepackage{tensor} 
\usepackage{cite}
\usepackage{bm}
\usepackage{threeparttable}
\usepackage{array}
\usepackage{multicol}  
\usepackage{multirow}
\usepackage{booktabs} 
\usepackage{verbatim}
\usepackage[utf8]{inputenc}
\usepackage{color, soul}
\usepackage[normalem]{ulem}

\theoremstyle{plain}
\newtheorem*{vimu_generation}{Virtual IMU Measurement}

\newcommand{\EqnEq} {\!\!\!\! & = & \!\!\!\!}

\newcommand{\Rot}{{\mathbf{R}}}
\newcommand{\Trans}{{\bm p}}
\newcommand{\Vel}{{\bm v}}
\soulregister\cite7
\soulregister\sout7
\soulregister\bf7

\title{
A Lightweight and Accurate Localization Algorithm Using Multiple Inertial Measurement Units 
}

\author{Ming Zhang, Xiangyu Xu, Yiming Chen, and Mingyang Li%
\thanks{Manuscript received: September, 10, 2019; Revised December, 17, 2019;
Accepted January, 13, 2020.}
\thanks{This paper was recommended for publication by Editor Eric Marchand upon
evaluation of the Associate Editor and Reviewers' comments.}
\thanks{The authors are with Alibaba Group, Hangzhou, China.
{\tt\small \{mingzhang, xiangyuxu, yimingchen, mingyangli\}@alibaba-inc.com.}}%
\thanks{Digital Object Identifier (DOI): see top of this page.}
}
\begin{document}

\maketitle

\markboth{IEEE Robotics and Automation Letters. Preprint Version. Accepted
January, 2020}
{ZHANG \MakeLowercase{\textit{et al.}}: A Lightweight and Accurate Localization Algorithm Using Multiple Inertial Measurement Units}
\begin{abstract}
This paper proposes a novel inertial-aided localization approach by fusing information from multiple inertial measurement units (IMUs) and exteroceptive sensors.
IMU is a low-cost motion sensor which provides measurements on angular velocity and gravity compensated linear acceleration of a moving platform, and widely used in modern localization systems. To date, most existing inertial-aided localization methods exploit only one single IMU. While the single-IMU localization yields acceptable accuracy and robustness for different use cases, the overall performance can be further improved by using {\em multiple} IMUs.
To this end, we propose a lightweight and accurate algorithm for fusing measurements from multiple IMUs and exteroceptive sensors, which is able to obtain noticeable performance gain without incurring additional computational cost. To achieve this, we first probabilistically map measurements from {\em all} IMUs onto a {\em virtual} IMU. This step is performed by stochastic estimation with least-square estimators and probabilistic marginalization of inter-IMU rotational accelerations. Subsequently, the propagation model for both state and error state of the virtual IMU is also derived, which enables the use of the classical filter-based or optimization-based sensor fusion algorithms for localization. Finally, results from both simulation and real-world tests are provided, which demonstrate that the proposed algorithm outperforms competing algorithms by noticeable margins.
\end{abstract}


\section{Introduction}
\IEEEPARstart{I}{n} recent years, commercial products which exploit inertial measurement units (IMUs) have been under fast development. This popular motion sensor can be found in robotics, personal electronic devices, wearable devices, and so on~\cite{koccer2013development}.
On one hand, the maturity of MEMS manufacturing process significantly reduces the size, price, and power consumption of the IMU hardware. On the other hand, significant progress has also been made in both algorithm and software design for IMUs, ranging from sensor characterization and calibration~\cite{flenniken2005characterization,li2014high, rehder2016extending, schneider2017visual}, measurement integration~\cite{trawny2005indirect, Li2013high,forster2017manifold}, sensor fusion\cite{qin2018vins,skog2016inertial,li2013real,schneider2018maplab,huai2018robocentric}, and so on.

In this work, we focus on the inertial-aided localization, 
which is to estimate the 6D poses (3D position and 3D orientation) of a moving platform. 
Since localization with only IMU inevitably suffers from pose drift, measurements from other sensors (i.e. aiding),
e.g., RGB cameras, depth cameras, or LiDARs (Light Detection And Ranging sensors), are typically used in combination with IMUs to provide long-term performance guarantees~\cite{schneider2018maplab,zhang2019localization,ye2019tightly}.
To perform accurate pose estimation, the majority of existing works use measurements from IMU for pose prediction, which is followed by probabilistic refinement using measurements from other sensors\cite{qin2018vins,skog2016inertial,li2013real,schneider2018maplab,Li2013high,forster2017manifold}. 

\begin{figure} 
	\centering 
	\subfigure{ 
		\includegraphics[width=\columnwidth]{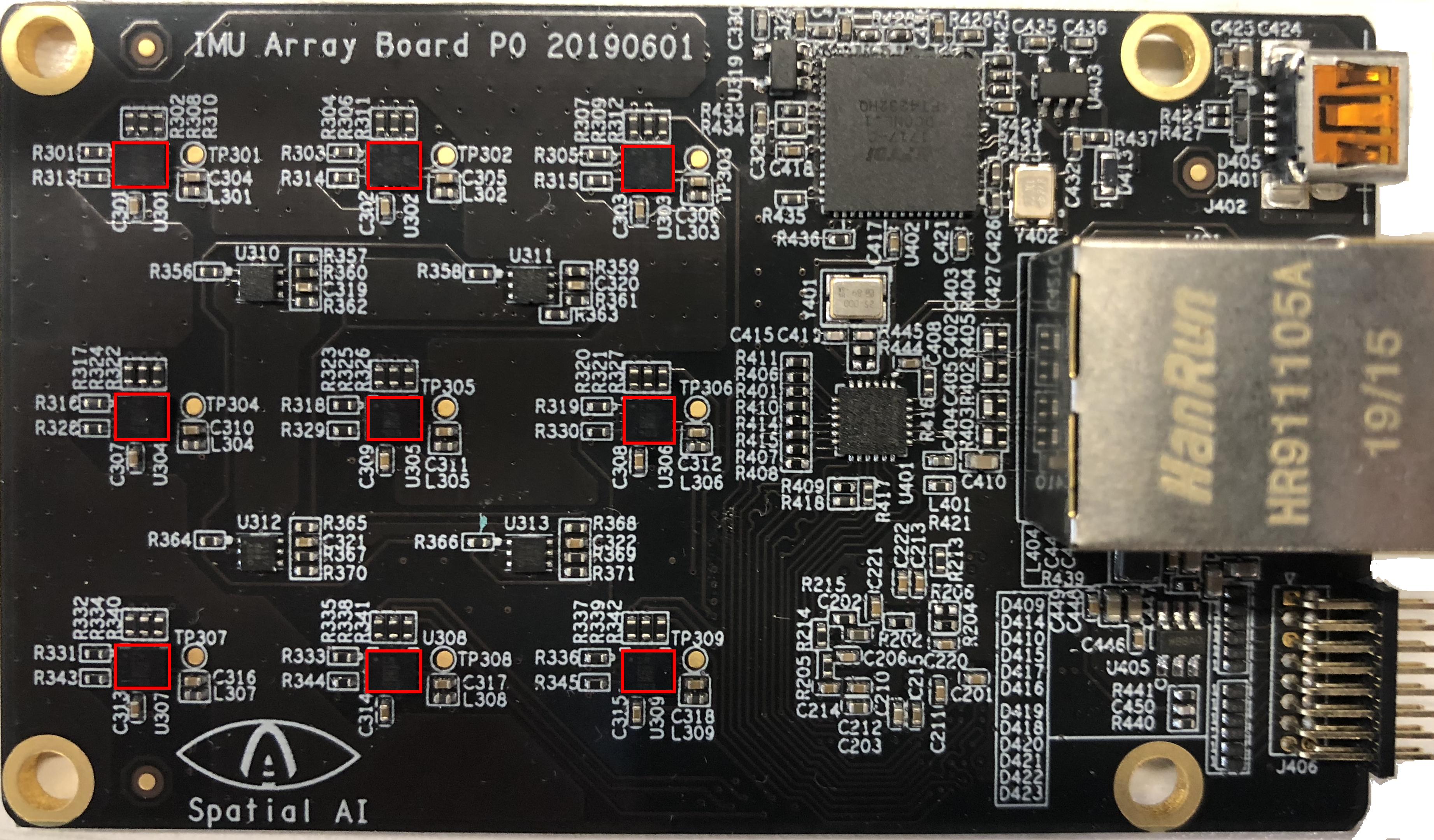} 
	}
	
	\caption{The IMU array board used in this work, which contains nine ST LSM6DSOX IMUs marked by red rectangles and a processor interface to connect cameras. The IMUs are synchronized by an embedded processor.} \label{fig:front}
\end{figure}

To date, most algorithms on inertial-aided localization are designed based on a {\em single} IMU\cite{qin2018vins,skog2016inertial,li2013real,schneider2018maplab,Li2013high,forster2017manifold,huai2018robocentric}. Although these algorithms are successfully deployed in different applications, using additional IMU sensors creates new possibilities for further improving the system accuracy and robustness.
Compared to other popular sensors for localization (e.g., cameras or LiDARs), IMUs especially the off-the-shelf MEMS ones are priced only hundredths or thousandths, and of smaller size as well as lower power consumption.
In addition, as a reliable proprioceptive sensor, IMU also poses less restrictions on operating environments and hardware configurations (in contrast, e.g. stereo cameras require enough spatial baseline to achieve performance gain~\cite{herath2006modeling,herath2007stereo}, which might not be feasible on various applications including mobile devices). 


%


Most existing methods on using multiple IMUs focus on processing IMU measurements only or integration with global navigation satellite systems (GNSS)~\cite{skog2016inertial,jafari2015optimal,shahri2017implementation}.
Fusing measurements from multiple IMUs with exteroceptive sensors for localization is a less-explored topic.
To the best of the authors' knowledge, the only work in recent years in this domain is~\cite{eckenhoff2019sensor}, which proposed an approach for vision-aided inertial navigation using measurements from multiple IMUs. However, the proposed algorithm is of significantly increased computation, which makes its real-world deployment on low-cost platforms infeasible.

In this paper, a systematic framework of using multiple IMUs is proposed for localizing a moving platform, in combination with exteroceptive sensors.
Compared to methods of using a single IMU, the proposed method is able to achieve better localization accuracy without incurring noticeable extra computation.
The key contribution of this work is the design of multiple-IMU propagation equation, which can be integrated into both
filter-based~\cite{li2013real,eckenhoff2019sensor} and optimization-based sensor fusion algorithms~\cite{leutenegger2015keyframe,ye2019tightly} for localization.	
To achieve this, we first probabilistically map measurements from all IMUs onto a \textit{virtual} IMU. This step is performed by stochastic estimation with least-square estimators and probabilistic marginalization of inter-IMU rotational accelerations.
Subsequently, the propagation models for both state and error state of the virtual IMU are derived, with closed-form representation of state-transition and noise Jacobian matrices.
By integrating them into a representative vision-aided inertial localization system, we show that the proposed method outperforms competing methods by a wide margin.
We also point out that, in this work, experimental analysis on localization accuracy with different number of IMUs is also provided, based on our customized sensor platform consisting of nine IMUs (see Fig.~\ref{fig:front}).

\section{Related Work} 
\label{sec:relatedwork}
We here group the related work by methods for inertial-aided localization and methods using multiple IMUs.

\subsection{Inertial Aided Localization}
IMU typically generates measurements at high frequencies and provides direct 6DoF (degrees of freedom) motion estimates of a moving platform. These properties make IMU an important complementary sensor for designing localization systems~\cite{Aided_INS}.

Camera is another widely used sensor in various applications. However, when used for localization, the methods only relying on cameras have theoretical drawbacks. They are unable to uniquely determine the roll and pitch angles against gravity as well as the metric scale (in monocular camera setup). This limits the localization accuracy and robustness. However, when an IMU is used together with a camera, these problems can be theoretically resolved, and the performance is largely improved. As a result, there are a variety of algorithms proposed in recently years on visual-inertial localization, 
ranging from estimator design~\cite{qin2018vins}\cite{leutenegger2015keyframe}, camera cost functions~\cite{zheng2017photometric}\cite{von2018direct}, 
efficiency and accuracy optimization~\cite{Li2013high}\cite{forster2017manifold}, sensor calibration~\cite{li2014high}\cite{rehder2016extending}, and so on.

IMU is also commonly used in combination with wheel encoders and/or laser range finders (LRFs) \cite{chen2019iros}. 
When combined with wheel encoders, IMU can help improve the dead reckoning and general localization precision~\cite{zhang2019localization}, as well as detect wheel slippery~\cite{yi2007imu}. On the other hand, LRF measures intensity and range from each laser beam at typically lower frequencies. Thus, the high-FPS measurements from an IMU can provide motion estimates between LRF scans and then derive more robust localization systems~\cite{ye2019tightly}.

\subsection{Methods of Using Multiple IMUs}
The methods of using multiple IMUs have a long history from inertial navigation community \cite{evans1970redundant}. However, the scope of early works is confined to fault detection and isolation (FDI), rather than localization accuracy until quite recently. An extensive literature review on multi-IMU FDI can be found in~\cite{yuksel2011optimal}. In what follows, we focus on the methods improving localization accuracy.
Early works on processing measurements from multiple IMUs aim on measurement noise reduction via least-square estimation~\cite{colomina2004redundant,waegli2008redundant}. However, 
the crucial correlation terms between accelerometer measurements and rotational accelerations are
ignored, and then the estimation becomes suboptimal. A method is proposed in \cite{yuksel2011optimal} to estimate acceleration/angular velocity based on a constrained optimization. These algorithms are later improved by~\cite{skog2016inertial}, which presents a two-stage method by 
firstly computing error-reduced rotational velocities and subsequently obtaining the rotational accelerations and specific force vectors.
\cite{herath2006modeling} focuses on improving calibration accuracy, designing an anti-parallel multi-IMU array and using IMUs of mixed measurement ranges to improve localization performance. Note that the virtual IMU is generated by simply averaging the data of multiple IMUs.
Another category of methods of using multiple IMUs is to design centralized estimators with a `stacked' state vector, concatenating the state of each IMU~\cite{bancroft2008twin,bancroft2009multiple}. This type of methods does not require approximations of sensors' measurements, and is typically of higher accuracy. However, the size of the concatenated state vector grows along with the number of IMU, and consequently so does the computation. 
The third type of methods is to design federated filters, using both local and master filters~\cite{carlson2001federated,bancroft2011data}. Shared states are estimated in both the local and master filter, while other states such as biases are only estimated in the local filters.

Note that all previously mentioned methods focus on IMU signal processing or GNSS localization only.
To the best of our knowledge, in existing literature, only~\cite{eckenhoff2019sensor} is dedicated to general localization with multiple IMUs and exteroceptive sensors. In \cite{eckenhoff2019sensor}, poses of each IMU are concatenated into the state vector, and relative pose constraints are derived between different IMUs. This approach is shown to generate results with higher accuracy, while at a cost of significantly increased computational complexity. In addition, although the general sensor fusion accuracy can be improved by~\cite{eckenhoff2019sensor}, the improvement will only take effect after each update. In other words, for applications that require high frequency pose estimates from IMU forward integration (e.g., real-time control for drones or low-latency rendering in virtual reality headsets),~\cite{eckenhoff2019sensor} has limitations. By contrast, our method is able to improve both overall localization  and forward integration accuracy without incurring additional computations.

The organization of the rest of this paper is as follows. Sec.~\ref{sec:notation} 
revisits the single-IMU localization method. 
Sec.~\ref{eqn:v_imu_meas} presents 
the virtual IMU model generated from multiple IMUs, which is used for deriving propagation equations in
Sec.~\ref{sec:prop}.
Finally, Sec.~\ref{sec:experiments} presents 
experimental results.

\section{Single IMU Method Revisited} 
\label{sec:notation}

This section revisits a standard method of using measurements from a single IMU for localization, which will be extended to the multiple-IMU case in the following sections.

\subsection{IMU Measurement Equations}
Assuming an IMU, $\{I\}$, moves with respect to a global frame $\{G\}$, the IMU measurements can be described by:
\begin{align}
 {\bm \omega}_m &\!=\! {^I}{\bm \omega} \!+\! {\bm b}_g \!+\! {\bm n}_g, \,\,\, {\bm n}_g \sim {\mathcal N}\left(\bm 0, {\bm \sigma}^2_{g} \mathbf I_3 \right), \label{eqn:w_m} \\
 {\bm a}_m &\!=\! {^I}{\bf R}{_G}({^G}{\bm a} \!-\! {^G}{\bm g}) \!+ \!{\bm b}_a \!+\! {\bm n}_a, \,\,\, {\bm n}_a \sim {\mathcal N}\left(\bm 0, {\bm \sigma}^2_{a} \mathbf I_3 \right) \label{eqn:a_m}
\end{align}
where ${\bm \omega}_m$ and ${\bm \alpha}_m \in \mathbb{R}^3$ are the gyroscope and accelerometer measurements, ${^I}{\bm \omega}$ and ${^G}{\bm a}$ the angular velocity and linear acceleration of the IMU expressed in frames $\{I\}$ and $\{G\}$ respectively, ${^I}{\bf R}{_G}$ the rotation from $\{G\}$ to $\{I\}$, ${\bm n}_g$ and ${\bm n}_a$ the white Gaussian noises, ${\bm b}_g$ and ${\bm b}_a$ the measurement biases modeled as random walk processes, and ${^G}{\bm g}$ the known gravity vector in global frame.

\subsection{IMU State Vector and Error State}
Conforming to~\cite{trawny2005indirect,Li2013high,schneider2018maplab}, the IMU state is defined as:
\begin{align}
\label{eq:state_single}
{\bm x} =
\left[
{\prescript{G}{I}{\bar{\bm q}}}^\intercal, ~
  {\bm b}^\intercal_{g}, ~
  {{^G}{\bm v}_I}^\intercal, ~
  {\bm b}^\intercal_{a}, ~
  {{^G}{\bm p}_I}^\intercal
  ~\right]^\intercal \in \mathbb{R}^{16}
\end{align}
where ${\prescript{G}{I}{\bar{\bm q}}} \in \mathbb{R}^4$ represents rotation from IMU frame to global frame (i.e. ${^G}\Rot_I$) in quaternion~\cite{trawny2005indirect}, and ${{^G}{\bm p}_I}$ and ${{^G}{\bm v}}_I$ stand for the IMU position and velocity. 
The continuous-time motion dynamics of the IMU are described as:
\begin{align}
\dot{\bm x} = f \left( 
\bm x,\,{\bm \omega}_m,{\bm a}_m
\right)
\end{align} 
whose detailed form is
\begin{align}
{^G}{\dot{\bm p}_I(t)}&\!=\!{^G {{\bm v}_I(t)}}, \;\;
\displaystyle {\dot{\bm b}_g}(t)\!=\!{\bm n}_{wg}(t), \;\;
\displaystyle {\dot{\bm b}_a}(t)\!=\!{\bm n}_{wa}(t)
, \\
{^G}{\dot{\bm v}_I(t)}&\!=\!{^G {\bm a}_I (t)}\!=\!
{^I}{\bf R}{_G^T} \left( {{\bm a}_m - 
	{\bm b}_a - {\bm n}_a}
\right) + {^G}{\bm g}
\\
{{^G_I}{}{\dot{\bar{\bm q}}(t)}}&\!=\!\frac{1}{2}\varOmega(^I{\bm  \omega}(t)){^G_I}{}\bar{\bm q}(t)\!=\!
\frac{1}{2}\varOmega(
{\bm \omega}_m \!-\! {\bm b}_g \!-\! {\bm n}_g
){^G_I}{}\bar{\bm q}(t)
\end{align}
where ${\bm n}_{wg}\sim{{\mathcal N}}({\bm 0}, {\bm \sigma}^2_{wg} \mathbf I_3)$, ${\bm n}_{wa}\sim{{\mathcal N}}({\bm 0}, {\bm \sigma}^2_{wa} \mathbf I_3)$,
\begin{align}
    \varOmega(^I \bm \omega) =  \begin{bmatrix}
   0 & -{^I \bm \omega}^\intercal \\
      ^I \bm \omega & -\lfloor ^I {\bm \omega}   \rfloor
\end{bmatrix},
\end{align}
and $\lfloor {\bm \cdot}   \rfloor$ represents the skew-symmetric matrix.
Similar to~\cite{trawny2005indirect,Li2013high,schneider2018maplab}, the IMU error state is defined as:
\begin{align}
\label{eq:err_single}
\tilde{\bm {x}} =
\left[
  {^I\tilde{ \bm \theta}}^\intercal, ~    
  {\tilde{\bm b}_g}^\intercal, ~
  {{^G}{\tilde{\bm v}_I}}^\intercal, ~
  {\tilde{\bm b}_a}^\intercal, ~
  {{^G}{\tilde{\bm p}_I}}^\intercal
\right]^\intercal  \in \mathbb{R}^{15}
\end{align}
in which the standard additive error definition is used for the position, velocity, and biases. The quaternion local error ${^I}\tilde{\bm \theta}$ is defined as:
\begin{equation}
{^I}\Rot_G\simeq (\mathbf I_3-\lfloor ^I \tilde{ \bm \theta}   \rfloor){^I}\hat{\Rot}_G
\end{equation}
where ${^I}\hat{\Rot}_G$ represents the estimate of ${^I}{\Rot}_G$.

\subsection{IMU Propagation Equations}
To implement probabilistic estimators, a general method is to derive a nonlinear equation for predicting the state estimate $\hat{\bm x}$ and a linearized one for characterizing the error state $\tilde{\bm x}$:
\begin{align}
\label{eq:est c}
\dot{\hat{\bm x}} &= f \left( 
\hat{\bm x},\,{\bm \omega}_m,{\bm a}_m
\right) \\
\label{eq:err c}
\dot{\tilde{\bm x}} &= 
\mathbf F \tilde{\bm x} + \mathbf G \mathbf n_c
\end{align} 
where $\mathbf F$ and $\mathbf G$ represent the continuous-time linearized state-transition matrix and noise Jacobian matrix, and $\mathbf n_c$ is the process noise. 
Once Eq.~\ref{eq:est c} and~\ref{eq:err c} are defined, IMU measurements can be straightforwardly integrated into different types of localization estimators for motion estimation~\cite{li2014high,trawny2005indirect, Li2013high,forster2017manifold,qin2018vins,li2013real,schneider2018maplab,zhang2019localization,ye2019tightly}.

\section{Virtual IMU Generation} \label{eqn:v_imu_meas}

Our high-level design principle of formulating multiple-IMU algorithm is to derive IMU propagation equations conforming to the structure of Eq.~\ref{eq:est c} and~\ref{eq:err c}, without non-probabilistic assumptions.
The main advantage of this design is that the proposed algorithm can be directly integrated into a variety of existing inertial-aided localization systems without any complicated software re-design.
To achieve this, we rely on information from multiple IMUs to generate measurements for a `virtual' IMU, which are used to replace ${\bm \omega}_m$ and ${\bm a}_m$ in Eq.~\ref{eq:est c}. The entire process is described in this section. In the next section, we derive the detailed equations for $\mathbf F$ and $\mathbf G$ in Eq.~\ref{eq:err c}, after the virtual IMU substitution.
We also note that, for the rest of this section, we discuss the representative case when two IMUs are involved, which can be straightforwardly extended to cases of more IMUs.

We begin by introducing our method of generating a `virtual' IMU from multiple real IMUs.
\begin{vimu_generation}
	Suppose that synchronized measurements ${\bm \omega}_{mA}$, ${\bm \omega}_{mB}$, ${\bm a}_{mA}$ and ${\bm a}_{mB}$ from IMU A and B are given, along with their extrinsic calibration $\left({^A}\Rot_B, {^B}\Trans_A \right)$. By arbitrarily picking the extrinsics $\left({^A}\Rot_V, {^V}\Trans_A \right)$ for the virtual IMU (reference frame $V$) relative to IMU A, the corresponding virtual gyroscope and accelerometer measurements can be generated as:
	\begin{eqnarray}
	{\bm \omega}_{mV} \EqnEq {\bf N}^+ {\bm \omega}_m \label{eqn:virtual_gyro} \\
	{\bm a}_{mV} \EqnEq {\bf T} \left({\bm a}_m - {\bf S}(\bm {\omega}_{mV})\right) \label{eqn:virtual_accel}
	\end{eqnarray}
	where
	\begin{align}
	\label{eq:def1}
	{\bf N} &\!\!=\!\!
	\begin{bmatrix}
	{^A}{\bf R}_V \\
	{^B}{\bf R}_V
	\end{bmatrix}, {\bm Y} \!\!=\!\! \begin{bmatrix}
	^A \mathbf{R}_V \lfloor{^V}{{\bm p}_A}  \rfloor\\
	^B \mathbf{R}_V \lfloor{^V}{{\bm p}_B}  \rfloor
	\end{bmatrix},
	{\bf T} \!\!=\!\! \left({\bf Z}^\intercal
    {\bf N}\right)^+ \! {\bf Z}^\intercal,
	\\
	\label{eq:def2}
	{\bm \omega}_m &\!=\!
	\begin{bmatrix}
	{\bm \omega}_{mA} \\
	{\bm \omega}_{mB}
	\end{bmatrix},
	{\bm a}_m \!=\! \begin{bmatrix}
	{\bm a}_{mA} \\
	{\bm a}_{mB}
	\end{bmatrix}\!,
	{\bf S} (\cdot)\! = \!
	\begin{bmatrix}
	{^A}{\Rot}_V{\lfloor{\cdot}\rfloor}^2 {^V}{{\bm p}}{_A}\\
	{^B}{\Rot}_V{\lfloor{\cdot}\rfloor}^2 {^V}{{\bm p}}{_B}
	\end{bmatrix},	
	\end{align}
	${\bf Z}^\intercal$ is left nullspace projection of
	${\bm Y}$, and ${\bf A}^+$ is the Moore-Penrose inverse of a real matrix ${\bf A}$, defined as:
	\begin{equation} \label{eqn:moore-penrose}
	{\bf A}^+ = \left({\bf A}^\intercal{\bf A}\right)^{-1}{\bf A}^\intercal.
	\end{equation}
\end{vimu_generation}
In practice, for general multiple IMU setup with ${\bf N} = \left[^1 \mathbf{R}_V^\intercal, ^2 \mathbf{R}_V^\intercal, \dots, ^{n_I} \mathbf{R}_V^\intercal \right]^\intercal$, the evaluation of ${\bf N}^+$ can be simplified as ${\bf N}^+ = {\bf N}^\intercal / n_I$.
With the generated virtual IMU measurements (Eq.~\ref{eqn:virtual_gyro} and~\ref{eqn:virtual_accel}), 
the prediction equation of the IMU state estimate (Eq.~\ref{eq:est c}) can be used without any modification.
The following two subsections discuss the methodology and derivations on generating the virtual IMU measurements.

\subsection{Virtual Gyroscope Model}
We first derive the measurement model for the virtual gyroscope.
To start with, we use the following identities:
\begin{equation} \label{eqn:virtual_gyro_to_real}
{^A}{\bm \omega} = {^A\Rot}_V {^V}{\bm \omega},\,
{^B}{\bm \omega} = {^B\Rot}_V {^V}{\bm \omega}.
\end{equation}
Combination of Eq.~\ref{eqn:w_m} and~\ref{eqn:virtual_gyro_to_real} leads to:
\begin{align}
\label{eq:stacked w}
\begin{bmatrix}
{\bm \omega}_{mA} \\
{\bm \omega}_{mB}
\end{bmatrix}&\!=\!
\begin{bmatrix}
{^A}{\bf R}_V \\
{^B}{\bf R}_V
\end{bmatrix}
{^V}{\bm \omega} \! + \!
\begin{bmatrix}
{\bm b}_{gA}\\
{\bm b}_{gB}
\end{bmatrix} \!+\!
\begin{bmatrix}
{\bm n}_{gA}\\
{\bm n}_{gB}
\end{bmatrix} 
\end{align}
Based on the measurements from IMUs $A$ and $B$, the optimal estimate of the virtual IMU angular velocity is given by:
\begin{align}
\label{eq:opt w}
{^V} \hat{\bm \omega} = \arg \min_{{\bm \omega} } \Bigg{|}\Bigg{|}
\begin{bmatrix}
\frac{{\bm \omega}_{mA}}{\sigma _{gA}} \\
\frac{{\bm \omega}_{mB}}{\sigma _{gB}}
\end{bmatrix}\!-\!
\begin{bmatrix}
\frac{1}{\sigma _{gA}}{^A}{\bf R}_V \\
\frac{1}{\sigma _{gB}}{^B}{\bf R}_V
\end{bmatrix}
{^V}{\bm \omega} \! - \!
\begin{bmatrix}
\frac{{\bm b}_{gA}}{\sigma _{gA}} \\
\frac{{\bm b}_{gB}}{\sigma _{gB}} 
\end{bmatrix} \Bigg{|}\Bigg{|}_2^2.
\end{align}
To simplify the analysis, we here assume $\sigma _{gA} = \sigma _{gB}$ for our derivations throughout the paper. Similar analysis can be developed for $\sigma _{gA} \neq \sigma _{gB}$. We also note that, if an IMU array is designed using the same type of IMUs, the $\sigma _{gA} = \sigma _{gB}$ assumption becomes exact. Solving for Eq.~\ref{eq:opt w} leads to the linear least-square estimate:
\begin{align}
\label{eq:virtual w est}
{^V} \hat{\bm \omega} 
= {{\bf N}^+ }
\begin{bmatrix}
{\bm \omega}_{mA} \\
{\bm \omega}_{mB}
\end{bmatrix} - 
{{\bf N}^+ }
\begin{bmatrix}
{\bm b}_{gA}\\
{\bm b}_{gB}
\end{bmatrix}
\end{align}
where ${{\bf N}^+ }$ is defined by Eq.~\ref{eq:def1} and \ref{eqn:moore-penrose}.
In addition, by substituting Eq.~\ref{eq:stacked w} into Eq.~\ref{eq:virtual w est}, we obtain:
\begin{align}
\label{eq:virtual w err}
{^V} {\bm \omega}   - {^V} \hat{\bm \omega}  = -
{{\bf N}^+ }
\begin{bmatrix}
{\bm n}_{gA}\\
{\bm n}_{gB}
\end{bmatrix}.
\end{align}
With Eq.~\ref{eq:virtual w est} and~\ref{eq:virtual w err} being defined, 
we are able to formulate the virtual gyroscope measurements, similar to the the original IMU measurements (i.e. Eq.~\ref{eqn:w_m}), as:
\begin{equation} 
\label{eqn:w_v_model}
{\bm \omega}_{mV} = {^V}{\bm \omega} + {\bm b}_{gV} + {\bm n}_{gV}, \,\,\, {\bm n}_{gV} \sim {\mathcal N}\left(\bm 0, {\bf Q}_{gV} \right)
\end{equation}
where ${\bm \omega}_{mV}$, ${\bm b}_{gV}$, and ${\bm n}_{gV}$ are the measurement, bias vector, and noise for the virtual IMU respectively, 
given by:
\begin{align}
\label{eq:w def}
{\bm \omega}_{mV} \!\!\triangleq\!\! {\bf N}^+
\begin{bmatrix}
{\bm \omega}_{mA} \\
{\bm \omega}_{mB}
\end{bmatrix},\,
{\bm b}_{gV} \!\!\triangleq \!\!{\bf N}^+
\begin{bmatrix}
{\bm b}_{gA} \\
{\bm b}_{gB}
\end{bmatrix},\,{\bm n}_{gV} \!\!\triangleq \!\!{\bf N}^+
\begin{bmatrix}
{\bm n}_{gA}\\
{\bm n}_{gB}
\end{bmatrix}.
\end{align}
We also note that, taking {\em expectation} on both sides of Eq.~\ref{eqn:w_v_model} results Eq.~\ref{eq:virtual w est}, and 
further calculating the error terms results in Eq.~\ref{eq:virtual w err}. The noise covariance matrix ${\bf Q}_{gV}$ in Eq.~\ref{eqn:w_v_model} 
can be computed as:
\begin{align}
\label{eq:QW}
{\bf Q}_{gV} \!\!=\!\! \sigma _{gA}^2 {{\bf N}^+ } {{{\bf N}^+ }}^T 
\!\!=\!\! \sigma _{gA}^2 \left( 
\begin{bmatrix}
{^A}{\bf R}_V \\
{^B}{\bf R}_V
\end{bmatrix}^T
\begin{bmatrix}
{^A}{\bf R}_V \\
{^B}{\bf R}_V
\end{bmatrix}
\right)^{-1}
\!\!=\!\!
\frac{\sigma _{gA}^2}{2} \mathbf I_3.
\end{align}
Eq. \ref{eq:QW} clearly indicates that, by using multiple IMUs, the overall IMU measurement noises can be reduced, and the localization accuracy should be improved.
When $\sigma _{gA} \neq \sigma _{gB}$, the expression in Eq.~\ref{eq:QW} becomes slightly different, but the conclusion on noise reduction still holds.

Additionally, based on Eq.~\ref{eq:w def}, the continuous time dynamics of ${\bm b}_{gV}$ can be described by:
\begin{equation}
\dot{\bm b}_{gV}(t)={\bm n}_{wgV}(t), \,\,\, {\bm n}_{wgV} \sim {\mathcal N}\left(\bm 0, {\bf Q}_{wgV} \right)
\end{equation}
and the above noises are modeled as 
\begin{align}
{\bf Q}_{wgV} & \triangleq {\bf N}^+
\begin{bmatrix}
{\sigma}^2_{wgA} \mathbf I_3& {\bf 0} \\
{\bf 0} & {\sigma}^2_{wgB}\mathbf I_3
\end{bmatrix} \left({\bf N}^+\right)^{\intercal},\label{eqn:q_bias_g} 
\end{align}
Similar to Eq.~\ref{eq:QW}, Eq.~\ref{eqn:q_bias_g} also indicates that by generating the virtual measurements, 
the corresponding gyroscope bias drifts slower.

\subsection{Virtual Accelerometer Model}
Similar to the virtual gyroscope, to derive the model 
for virtual accelerometer, we start with the identity:
\begin{align} \label{eq:g_p_a}
{^G}{\bm p}_{A} = {^G}{\bm p}_{V} + {^G}\Rot_{V}{^V}{\bm p}_{A}.
\end{align}
Taking the first and second derivative of Eq. \ref{eq:g_p_a} leads to:
\begin{align}
{^G}\Vel _{A} &\!=\! {}
{^G}\Vel _{V}  \!+ \!{{^G}\Rot_{V}} \lfloor {}^V \boldsymbol{\omega} \rfloor
{{^V}{\bm p}_{A} }, \nonumber \\
{^G}{\bm a}  _{A} &\!=\! {}
{^G}{\bm a} _{V}
\!+\!{^G}\Rot_{V} 
{\lfloor {}^V \boldsymbol{\omega} \rfloor}^2  {}
{{^V}{\bm p}_{A} } \!+ \!
{}{^G}\Rot_{V}  \lfloor 
{}^V \boldsymbol{\phi} \rfloor 
{}{^V}{\bm p}_{A} \label{eqn:G_a_A}
\end{align}
where ${^V}{\bm \omega}, {{^V}\boldsymbol{\phi}} \in \mathbb{R}^3$ are the angular velocity and the angular acceleration of virtual IMU frame in global frame.

Substituting Eq.~\ref{eqn:a_m} into Eq.~\ref{eqn:G_a_A} leads to
\begin{align}
\label{eq:a before}
{\bm a}_{mA}-\mathbf b_{aA}-{\bm n}_{aA} = {^A}\Rot_{V} \left({\bm s} _{V}
\!+\!
{\lfloor {}^V \boldsymbol{\omega} \rfloor}^2  {}
{{^V}{\bm p}_{A} } \!- \!
\lfloor {}{^V}{\bm p}_{A} 
\rfloor 
{}^V \boldsymbol{\phi}  \right)
\end{align}
where the identity $\bm a \lfloor {\bm b} \rfloor = -\bm b \lfloor {\bm a} \rfloor$ is applied, and
\begin{equation} \label{eqn:specific_force}
{\bm s_V} \triangleq {}{^V}\Rot_{G} ({^G}{\bm a}_{V} - {^G}{\bm g})
\end{equation}
represents the specific force vector \cite{Aided_INS}.

Similar to computing the optimal estimate of $\bm \omega$ in Eq.~\ref{eq:opt w}, stacking accelerometer measurements from both IMUs
via Eq.~\ref{eq:a before} leads to:
\begin{align}
\label{eq:opt a}
\hat{\bm s}_{V} \!\!=\!\! \arg \min_{{\bm s}_{V}  } \!\!\Bigg{|}\Bigg{|}\!\!
\begin{bmatrix}
{\bm a}_{mA} \\
{\bm a}_{mB}
\end{bmatrix}\!\!-\!\!
\begin{bmatrix}
{^A}{\bf R}_V \\
{^B}{\bf R}_V
\end{bmatrix} \!\!
{\bm s}_{V} \! \!+ \!\!
{\bf Y}{}{^V}{\bm \phi} \!\!-\!\! {\bf S}({^V}{\bm \omega}) \!\!- \!\!
\begin{bmatrix}
{\bm b}_{aA} \\
{\bm b}_{aB}
\end{bmatrix}\!\! \Bigg{|}\Bigg{|}_2^2\!\!
\end{align}
where matrix $\bf Y$ and operator $\bf S (\cdot)$ are defined in Eq.~\ref{eq:def1} and~\ref{eq:def2} respectively. 
In Eq.~\ref{eq:opt a}, the rotational acceleration ${^V}{\bm \phi}$ is {\em unknown}, and also {\em not} directly measured by 
the IMU sensor. If ${^V}{\bm \phi}$ is not properly handled, ${\bm s}_{V}$ can not be accurately represented. 
In this work, we adopt a method inspired by~\cite{Li2013high}, in which {\em unknown} visual features are probabilistically marginalized. 
By denoting $\bf Z^T$ the left nullspace projection \footnote{$\bf Z^T$ can be computed through QR decomposition \cite{Li2013high}.} of the matrix $\bf Y$, Eq.~\ref{eq:opt a} is equivalent to:
\begin{align}
\label{eq:opt a1}
\hat{\bm s}_{V} \!\!=\!\! \arg \min_{{\bm s}_{V}  } \!\!\Bigg{|}\Bigg{|}
\bf Z^T \left( \begin{bmatrix}
{\bm a}_{mA} \\
{\bm a}_{mB}
\end{bmatrix}\!\!-\!\!
\begin{bmatrix}
{^A}{\bf R}_V \\
{^B}{\bf R}_V
\end{bmatrix} \!\!
{\bm s}_{V} \! \!- \!\! {\bf S}({^V}{\bm \omega}) \!\!- \!\!
\begin{bmatrix}
{\bm b}_{aA} \\
{\bm b}_{aB}
\end{bmatrix} \right) \!\! \Bigg{|}\Bigg{|}_2^2\!\!
\end{align}
Solving Eq. \ref{eq:opt a1}, it follows that
\begin{align} 
\label{eq:sv}
\hat{\bm s}_V= {\bf T} \left(\begin{bmatrix}
{\bm a}_{mA} \\
{\bm a}_{mB}
\end{bmatrix}-{\bf S}({^V}{\bm \omega})-\begin{bmatrix}
{\bm b}_{aA} \\
{\bm b}_{aB}
\end{bmatrix}\right),
\end{align}
with $\bf T$ also defined in Eq.~\ref{eq:def1}.
Eq.~\ref{eq:sv} allows us to define the virtual accelerometer measurement as Eq.~\ref{eqn:virtual_accel}.
Combining Eq.~\ref{eqn:virtual_accel}, ~\ref{eq:a before} and~\ref{eq:sv}, we are able to write:
\begin{align} 
\label{eqn:a_v_model}
{\bm a}_{mV} = {\bm s}_V + {\bm b}_{aV} + {\bm n}_{aV} - 
{\bf T} \cdot {\bf S}_{\bm a}
\end{align}
with
\begin{align}
\label{eq:a def}
{\bm b}_{aV} \triangleq {\bf T}
\begin{bmatrix}
{\bm b}_{aA} \\
{\bm b}_{aB}
\end{bmatrix},\,  
{\bm n}_{aV} \triangleq {\bf T} 
\begin{bmatrix}
{\bm n}_{aA}\\
{\bm n}_{aB}
\end{bmatrix}
\end{align}
and
\begin{align}
{\bf S}_{\bm a} = 
{\bf S} ({\bm \omega}_{mV} \!\!-\!\! {^V{\bm \omega}})
 = \begin{bmatrix}
{^A}{\Rot}_V {\bm \zeta}  {^V}{{\bm p}}{_A}\\
{^B}{\Rot}_V {\bm \zeta}  {^V}{{\bm p}}{_B}
\end{bmatrix},\,{\bm \zeta} \!\!=\!\! {\lfloor  \bm {\omega}_{mV}  \rfloor}^2 \!\!-\!\!{\lfloor  ^V\bm {\omega}  \rfloor}^2
\end{align}
where ${\bm \zeta} $ can be expanded as
\begin{align}
{\bm \zeta} & = {\lfloor  \bm {\omega}_{mV}  \rfloor}^2 -
{\lfloor  \bm {\omega}_{mV} -  {\bm {b}_{gV}} - {\bm {n}_{gV}} \rfloor}^2 \\
&\simeq\!-\!{\lfloor\bm {b}_{gV} \rfloor}^2
\!\!+\!
{\lfloor\bm {\omega}_{mV}  \rfloor}
{\lfloor\bm {b}_{gV} \rfloor}\!+\!
{\lfloor\bm {n}_{gV} \rfloor}
{\lfloor\bm {\omega}_{mV}  \rfloor}\!-\!
{\lfloor\bm {b}_{gV} \rfloor}
{\lfloor\bm {\ n}_{gV} \rfloor} \notag \\
&+{\lfloor\bm {b}_{gV} \rfloor}
{\lfloor\bm {\omega}_{mV}  \rfloor}\!+\!
{\lfloor\bm {\omega}_{mV}  \rfloor}
{\lfloor\bm {n}_{gV} \rfloor}\!-\!
{\lfloor\bm {\ n}_{gV} \rfloor}
{\lfloor\bm {b}_{gV} \rfloor}.
\label{eq:zeta}
\end{align}
In Eq. \ref{eq:zeta}, we have used the quadratic error approximation ${\lfloor\bm {n}_{gV} \rfloor}^2 \simeq \mathbf 0_3$.

Eq.~\ref{eqn:a_v_model} -~\ref{eq:zeta} clearly demonstrate that, the virtual accelerometer measurements are affected by gyroscope biases, accelerometer biases, gyroscope noises, and accelerometer noises from original IMUs. 
This is unlike the original IMU measurement equations, i.e., Eq.~\ref{eqn:w_m} and~\ref{eqn:a_m}, in which measurement noises 
are {\em not} correlated and biases are independent.


\section{Virtual IMU Propagation}
\label{sec:prop}

In this section, a method is presented to integrate virtual IMU measurements, i.e. Eq.~\ref{eqn:virtual_gyro} and~\ref{eqn:virtual_accel}, for both the IMU state and error state.
Similar to the case of single IMU, i.e. Eq.~\ref{eq:state_single} and~\ref{eq:err_single}, we define the state and error state of the virtual IMU as:
\begin{align}
\label{eq:state_v}
{\bm x}_V &=
\left[
{\prescript{G}{V}{\bar{\bm q}}}^\intercal, ~
{\bm b}^\intercal_{gV}, ~
{{^G}{\bm v}_V}^\intercal, ~
{\bm b}^\intercal_{aV}, ~
{{^G}{\bm p}_V}^\intercal
\right]^\intercal \in \mathbb{R}^{16} \\
\label{eq:err_v}
\tilde{\bm {x}}_V &=
\left[
{^V\tilde{ \bm \theta}}^\intercal, ~
{\tilde{\bm b}_{gV}}^\intercal, ~
{{^G}{\tilde{\bm v}_V}}^\intercal, ~
{\tilde{\bm b}_{aV}}^\intercal, ~
{{^G}{\tilde{\bm p}_V}}^\intercal
\right]^\intercal  \in \mathbb{R}^{15}.
\end{align}
whose continuous-time propagation equations are given by
\begin{align}
\label{eq:est v}
\dot{\hat{\bm x}}_V &= f \left( 
\hat{\bm x},\,{\bm \omega}_{mV},{\bm a}_{mV}
\right) \\
\label{eq:err v}
\dot{\tilde{\bm x}}_V &= 
\mathbf F_V \tilde{\bm x}_V + \mathbf G_V \mathbf n_V
\end{align} 
with
\begin{align}
\mathbf n_V &=\left[
{\bm n_{gV}}^\intercal, ~
{\bm n_{wgV}}^\intercal, ~
{\bm n_{aV}}^\intercal, ~ 
{\bm n_{waV}}^\intercal~
\right]^\intercal \in \mathbb{R}^{12}.
\end{align}
Eq. {\ref{eq:est v}} and {\ref{eq:err v}} provide core equations for performing state and error state propagation based on the virtual IMU. This propagation can be straightforwardly integrated into different types of tightly-coupled localization algorithms, identical to the process of using a single IMU. To implement Eq.~\ref{eq:est v}, we first generate the virtual IMU measurements from all IMUs via Eq.~\ref{eqn:virtual_gyro} and~\ref{eqn:virtual_accel}, and the remaining step of pose integration is identical to that of Eq.~\ref{eq:est c}.
By denoting 
\begin{align}
\hat{\bm \omega} &= {\bm \omega}_{Vm} - \hat{\mathbf b}_{gV},\,\,
\hat{\bm a}_V = {\bm a}_{Vm} - \hat{\mathbf b}_{aV} + 
\mathbf T \hat{\bf S}_a
\end{align}
the state-transition matrix and noise Jacobian matrix in Eq.~\ref{eq:err v} can be computed, and the detailed expression is given by Eq.~\ref{eq:big}.
\begin{figure*}[htbp]
\begin{align}
\label{eq:big}
\bf F_V=  \begin{bmatrix}
     -\lfloor \hat{\bm \omega}   \rfloor&
     -\bf I_{3}& \bf 0_{3}& 
     \bf 0_{3}& \bf 0_{3}\\
     \bf 0_{3}& \bf 0_{3}& 
     \bf 0_{3}& \bf 0_{3}&
     \bf 0_{3} \\
     {-^V}\hat{\bf R}_G^\intercal \lfloor \hat{\bf a}_V   \rfloor & {-^V}\hat{\bf R}_G^\intercal\bf T {\bf \Psi}& \bf 0_{3}& {-^V}\hat{\bf R}_G^\intercal& \bf 0_{3}& \\
     \bf 0_{3}& \bf 0_{3}& 
     \bf 0_{3}& \bf 0_{3}&
     \bf 0_{3} \\
     \bf 0_{3}& \bf 0_{3}& 
     \bf I_{3}& \bf 0_{3}&
     \bf 0_{3}
    \end{bmatrix},
        \bf G_V=\begin{bmatrix}
-\bf I_{3}& \bf 0_{3}& 
     \bf 0_{3}& \bf 0_{3}&\\
 \bf 0_{3}& \bf I_{3}& 
     \bf 0_{3}& \bf 0_{3}&\\
     \bf 0_{3}& {-^V}\hat{\bf R}_G^\intercal\bf T  {\bf \Xi}&
     {-^V}\hat{\bf R}_G^\intercal& \bf 0_{3}\\
     \bf 0_{3}& \bf 0_{3}& 
     \bf 0_{3}& \bf I_{3}\\
     \bf 0_{3}& \bf 0_{3}& 
     \bf 0_{3}& \bf 0_{3}
\end{bmatrix} 
\end{align}
\end{figure*}
 Compared to the single IMU case, $\mathbf F_V$ and $\mathbf G_V$ contain the following additional non-zero terms:
 \begin{align}
 \frac{\partial {^G}{\dot{\tilde{\bm v}}_V}}{\partial \tilde{\bm b}_{gV}} 
 =
 {-^V}\hat{\bf R}_G^\intercal{\bf T} {\bf \Psi},\,\,\,
 \frac{\partial {^G}{\dot{\tilde{\bm v}}_V}}{\partial {\bm n}_{gV}} 
 =
 {-^V}\hat{\bf R}_G^\intercal\bf T {\bf \Xi}
 \end{align}
where
\begin{align}
{\bm \Psi} = {\bm \Xi} = 
\begin{bmatrix}
{^A}{\Rot}_V \left( -
 {\lfloor{}^V \hat{\bm {\omega}}  \rfloor}
{\lfloor {{^V}{{\bm p}}{_A}}\rfloor} -
\lfloor 
{\lfloor{}^V \hat{\bm {\omega}}  \rfloor} {^V}{{\bm p}}{_A} \rfloor \right)  \\
{^B}{\Rot}_V \left( -
 {\lfloor{}^V \hat{\bm {\omega}}  \rfloor}
{\lfloor {{^V}{{\bm p}}{_B}}\rfloor} -
\lfloor 
{\lfloor{}^V \hat{\bm {\omega}}  \rfloor} {^V}{{\bm p}}{_B} \rfloor \right) 
\end{bmatrix}.
\end{align}
The detailed derivations of ${\bm \Psi}$ and ${\bm \Xi}$ are shown in Appendix. 
In addition to those two terms, the expressions of all other components of $\mathbf F_V$ and $\mathbf G_V$ are identical to those in the single IMU case.
Once detailed expression of both matrices in Eq.~\ref{eq:err v} is given, a computer programmable discrete-time estimator can be derived accordingly, similar to existing work on IMU integration~\cite{trawny2005indirect,forster2017manifold,schneider2018maplab}. 

It is also important to note that compared to processing measurements of a single IMU, 
the additional computation introduced by the proposed algorithm can be neglected. The major additional computational operations include i) computing nullspace projection of matrix $\bf Y$ and obtaining $\bf T$ in Eq.~\ref{eq:def1} and ii) computing discrete time error state propagation by handling extra fill-in in $\mathbf F_V$ and $\mathbf G_V$. Note that the computational complexity of the first task is \textit{linear} in the number of IMUs while can be done offline given the IMUs extrinsics. The added computational cost of the second task is constant. By contrast, the computational cost of the method in~\cite{eckenhoff2019sensor} is cubic in the total number of camera poses and IMUs, and thus adding IMUs incurs significant extra computing time. This is not the case in our proposed method.

\begin{figure*}[t]
	\centering
	\includegraphics[scale=0.4]{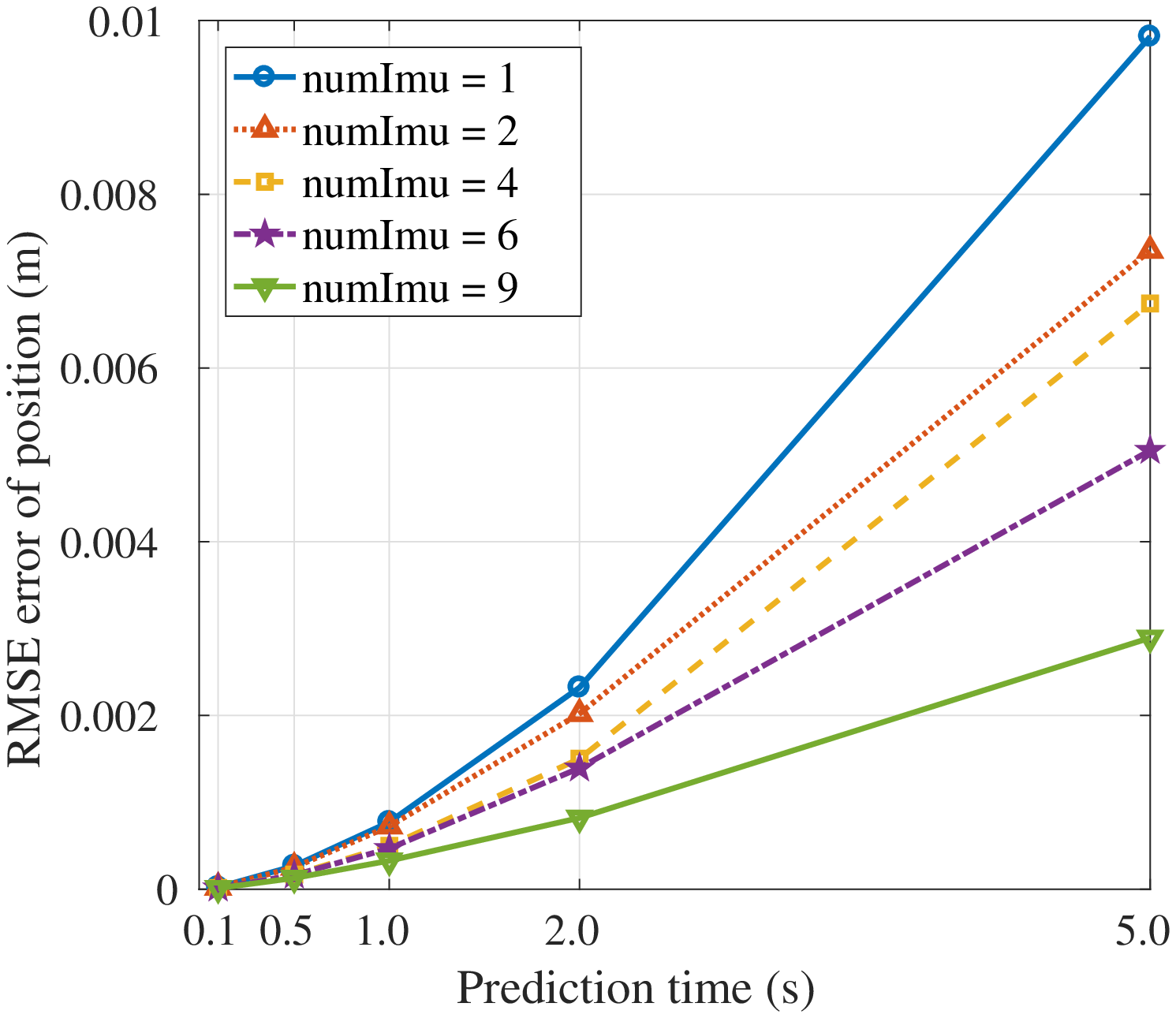}
	\includegraphics[scale=0.4]{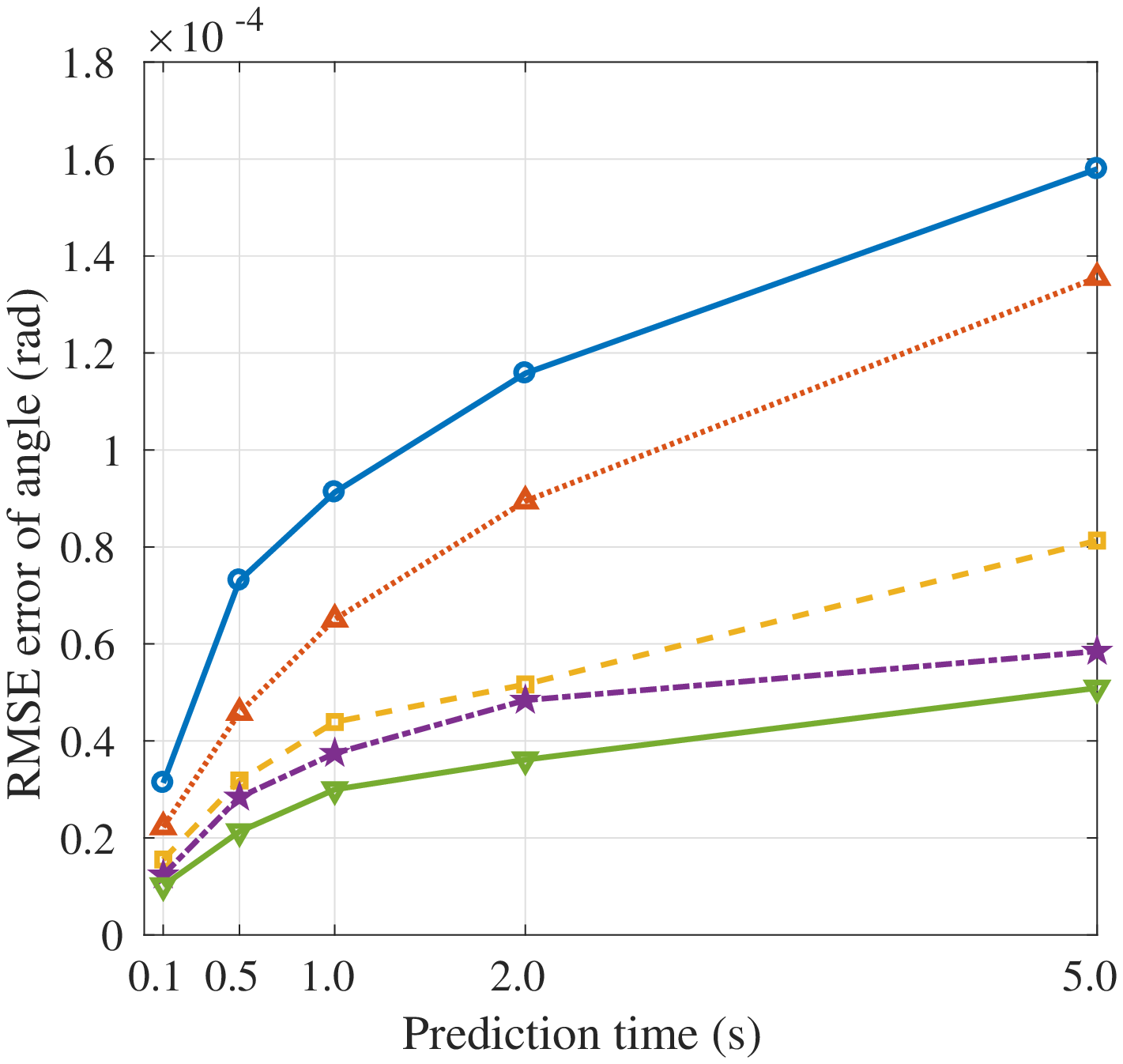}
	\includegraphics[scale=0.4]{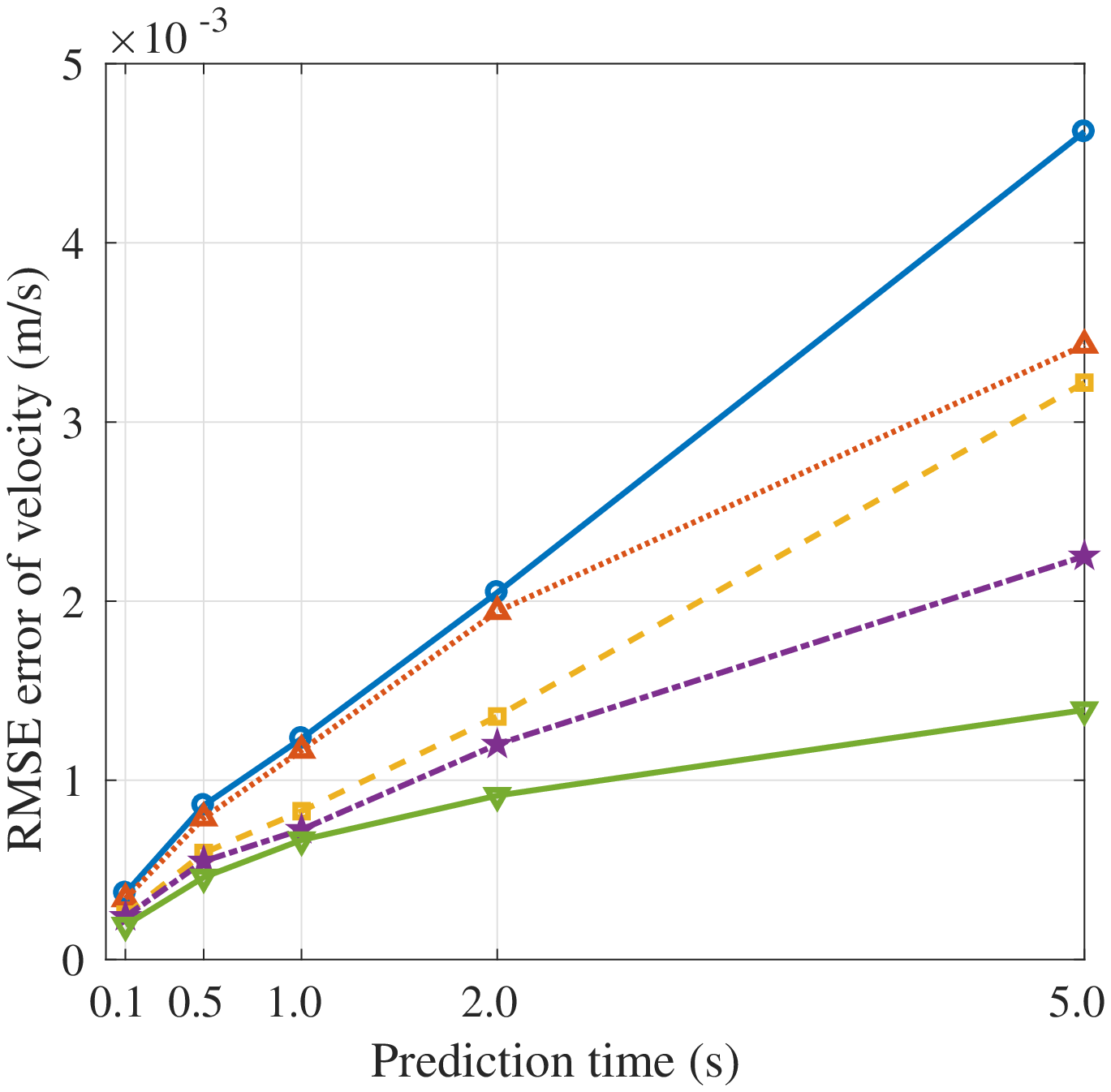}
	\caption{IMU integration errors when different number of IMUs used for varying time durations. Left: Errors in predicted position; Middle: Errors in predicted orientation; Right: Errors in predicted velocity.}
	\label{fig:prediction_error}
\end{figure*}
\section{EXPERIMENTS} \label{sec:experiments}

\subsection{Methodology}
To demonstrate the performance of our proposed localization algorithm with multiple IMUs, we integrate it into a visual-inertial odometry (VIO) approach~\cite{zhang2019localization}\footnote{In our implementation of~\cite{zhang2019localization}, only measurements from the IMUs and cameras are used.}. The performance of VIO is tested against different number of IMUs. When there is only one IMU used, the VIO algorithm is identical to~\cite{zhang2019localization}. When more IMUs are used, the proposed method of generating virtual IMU measurements and performing virtual IMU propagation replaces the single-IMU pipeline. In addition, the method of~\cite{eckenhoff2019sensor} is implemented in our tests, as a competing algorithm to compare against.

\subsection{Simulation Tests}
We first show results from simulation tests. To minimize the reality gap of simulation, 
we generated synthetic poses and sensor measurements based on real-world data, similar to~\cite{li2014vision}.
In the tests, we used $9$ IMUs and a monocular camera, with perfectly known sensor intrinsic and extrinsic parameters. During data generation, each IMU was sampled at $200$Hz and the camera captured measurements at $10$Hz. During the simulation, the layout of the multiple IMU array is identical to that of Fig.~\ref{fig:front}, and 
all IMUs were synchronized.

\subsubsection{Pose Prediction Error}
The first test is to demonstrate the pose integration accuracy by using different number of IMUs for varying time durations. Specifically, we started this test with {\em perfectly} known IMU pose, i.e., position, orientation, velocity, biases, and used sensor measurements to predict poses in future timestamps. For all methods, we computed the root-mean-square (RMS) errors for 3D position, rotation and velocity over 50000 simulation tests. 
Specifically, we generated 50 data sequences, and sampled 1000 different IMU poses from each sequence for performing prediction. We also note that~\cite{eckenhoff2019sensor} is identical to the single IMU case in this test, since the formulation in~\cite{eckenhoff2019sensor} is not able to improve the IMU prediction accuracy.

\begin{figure*}[t]
\centering
\includegraphics[width=.45\textwidth]{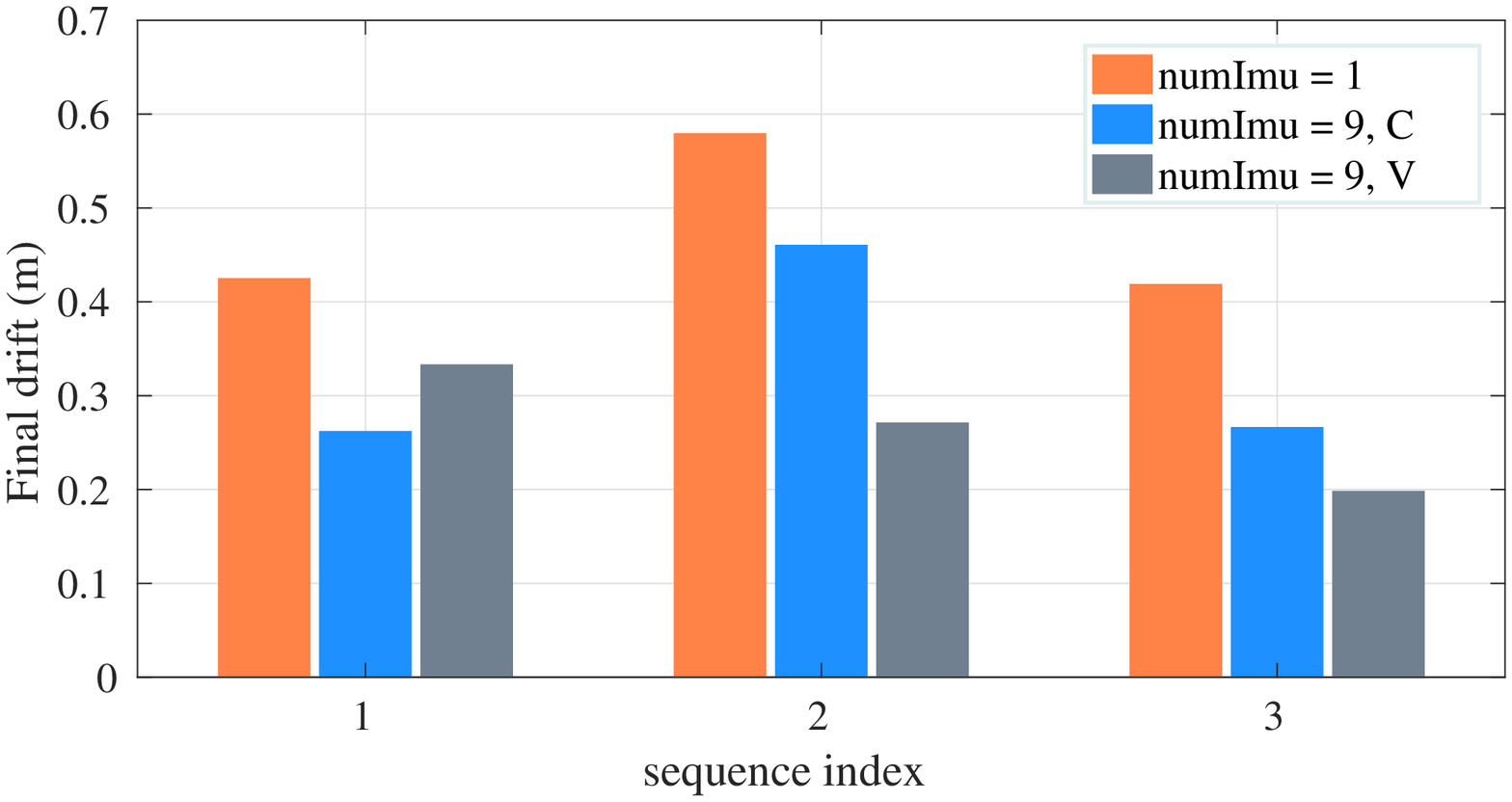}
\includegraphics[width=.45\textwidth]{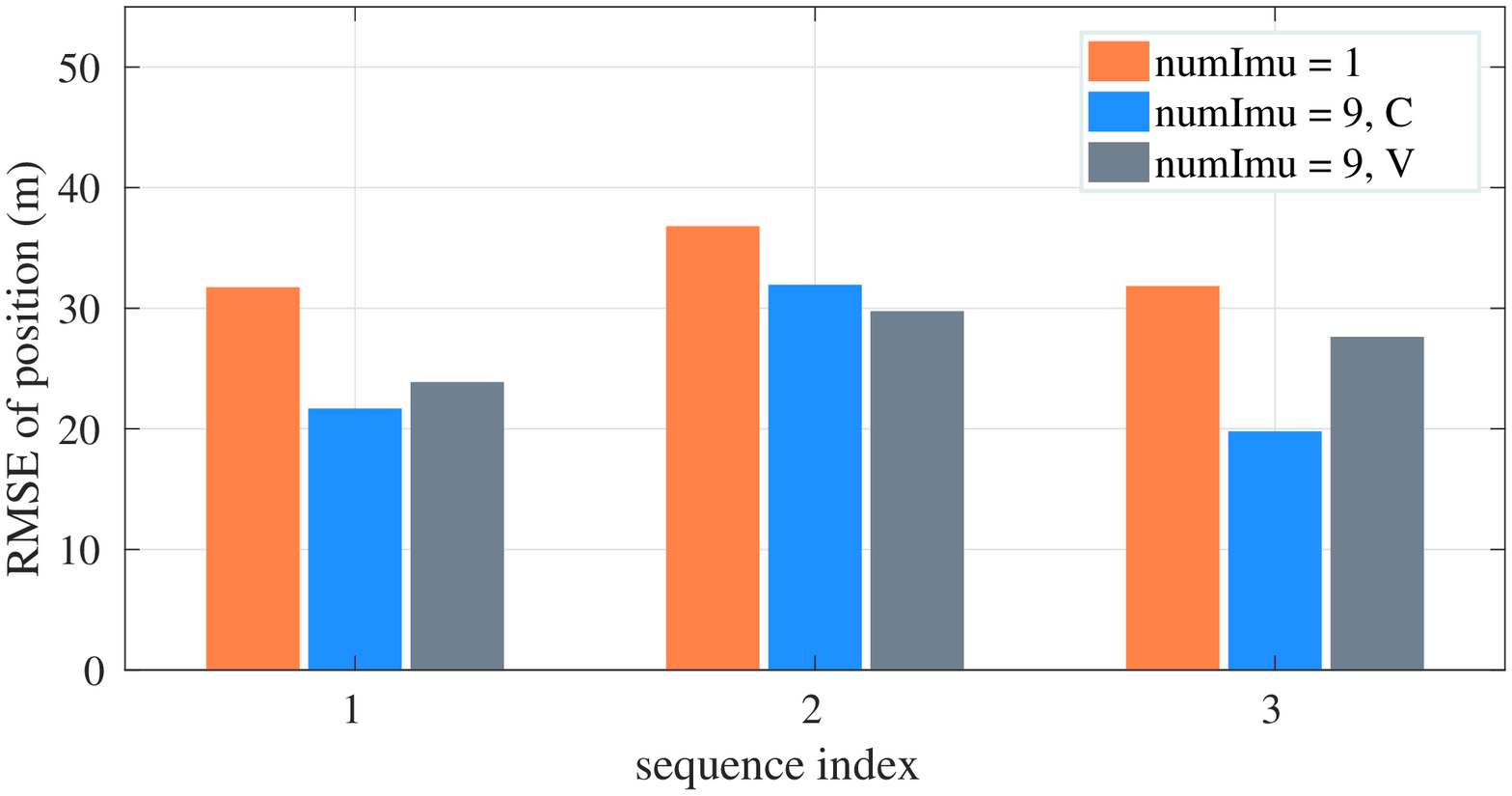}
\caption{Localization errors in real-world experiments. Three methods are tested: The original VIO algorithm using a single IMU, the modified VIO algorithm using the proposed method with $9$ IMUs (method V, V = Virtual), and~\cite{eckenhoff2019sensor} with 9 IMUs (method C, C = Centralized). The left plot represents the final drifts in indoor tests, and the right plot denotes the positional RMS errors in urban street tests.}
\label{fig:exp error}
\end{figure*}

 \begin{table}
	\centering  
	\begin{threeparttable}  
		\caption{Pose RMS errors as a function of number of IMUs.}  \label{table: pose_error}
		\begin{tabular}{l p{0.74cm}<{\centering}
				p{0.74cm}<{\centering}
				p{0.74cm}<{\centering}
				p{0.74cm}<{\centering} 
				p{0.74cm}<{\centering} 
			}  
			
			\toprule  
			\quad Num. of IMUs & 1 & 2 & 4 & 6 & 9 \\

			\midrule
			\multicolumn{1}{l}{\bf{Environment 1}} \cr
			\quad Pos. err. (m)    & 0.1985  & 0.1853 &  0.1572 &  0.1524 & \bf{0.1455} \\
			\quad Rot. err. (rad.) & 0.0041  & 0.0040 &  0.0036 &  0.0035 & \bf{0.0031} \\
			
			\midrule
			\multicolumn{1}{l}{\bf{Environment 2}} \cr
			\quad Pos. err. (m)    & 0.3917 & 0.3721 & 0.3117 &  0.3105 & \bf{0.2955} \\
			\quad Rot. err. (rad.) & 0.0054 & 0.0052 & 0.0048 &  0.0048 & \bf{0.0041} \\
			
			\bottomrule  
		\end{tabular}  
	\end{threeparttable}  
\end{table} 

The testing results are shown in Fig.~\ref{fig:prediction_error}, which
clearly demonstrate that by using more IMUs for pose integration the accuracy can be largely improved. This validates our theoretical analysis, and our design motivation of using more IMUs. In addition, we emphasize that, in all cases, using multiple IMUs for pose prediction is consistently better than the original method of using a single IMU and that of~\cite{eckenhoff2019sensor}.

\subsubsection{VIO Localization Error}
The second test is to demonstrate 
the localization accuracy when the proposed method is integrated into VIO. Similar to the previous tests, we conducted statistical comparison between the cases when different number of IMUs were involved. For all methods, we computed the RMS errors for both 3D position and rotation, under two sets of representative simulation environments.

Table~\ref{table: pose_error} 
shows the results when $1$, $2$, $4$, $6$, and $9$ IMUs are used. The most important conclusion from Table~\ref{table: pose_error} is that when multiple IMUs are used via the proposed method the localization accuracy is consistently higher than that of the single-IMU based method. In addition, we also observe that when more IMUs are used, the accuracy can be further improved, in terms of both rotational and positional estimates.

\subsection{Real World Experiments}
\subsubsection{Testing Platforms and Localization Environment}
\begin{figure} 
	\centering 
	\subfigure{ 
		\includegraphics[width=\columnwidth,height=5cm]{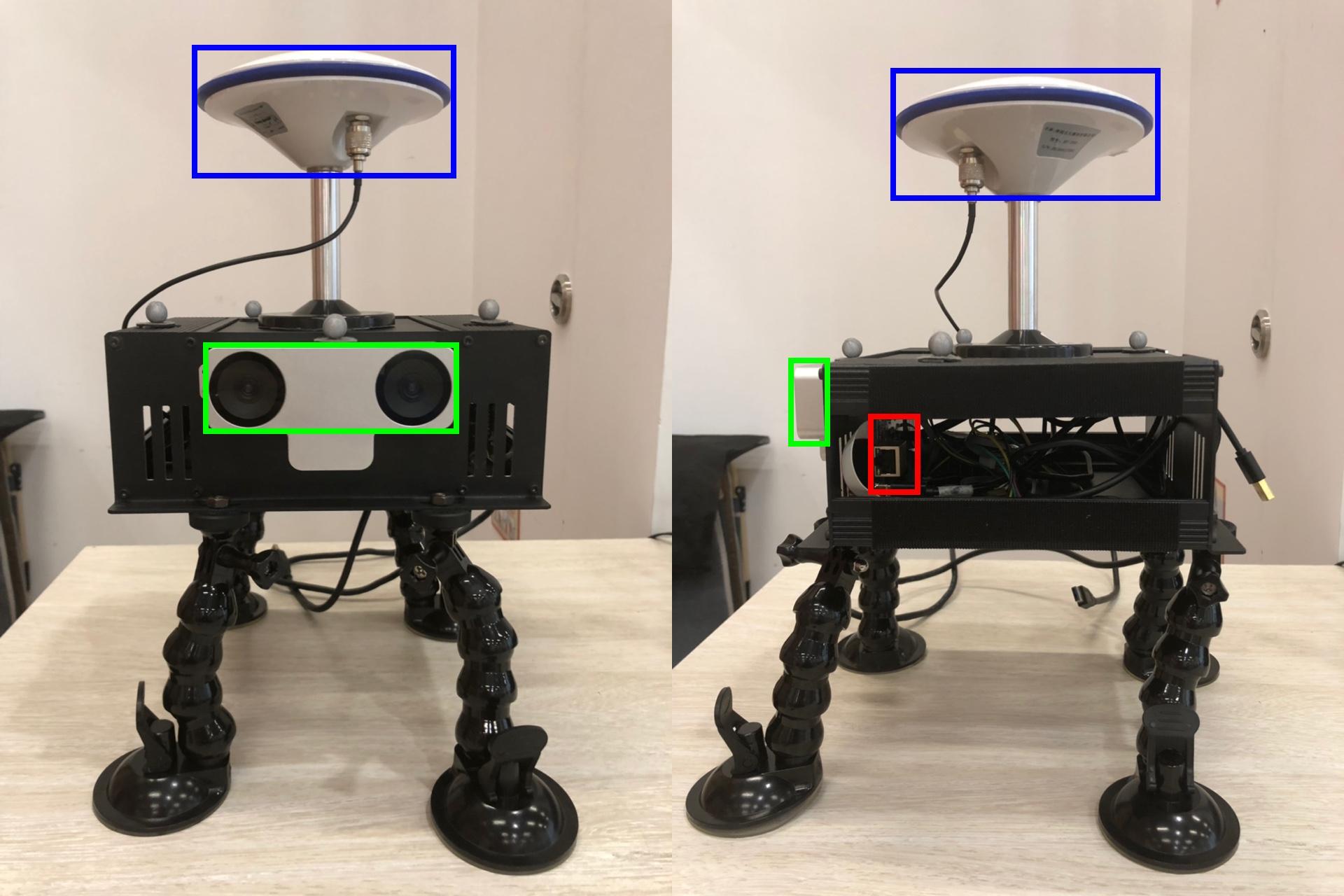} 
	}
	\caption{Sensor platform in our tests. GPS antenna is marked by a blue rectangle, the stereo camera system by a green one and the IMU array board by a red one. Left figure: The front view. Right figure: The side view.} \label{fig:sensors}
\end{figure}

To evaluate the performance of the proposed method, we also conducted experiments by using data sets from our customized sensor platform.
Fig. {\ref{fig:sensors}} shows that our sensor platform consists of a stereo camera system with ON AR0144 imaging sensors, an array of $9$ ST LSM6DSOX IMUs (see Fig.~{\ref{fig:front}}) and a RTK-GPS system. In our experiment, we chose the virtual IMU frame identical to the central IMU in the IMU array.

In our experiment, the multiple IMUs are synchronized by hardware design using IMUs' interrupt signals~\cite{li2004cost}, and performed extrinsic calibration between IMUs offline via~\cite{rehder2016extending}. The camera intrinsics and camera to IMU extrinsics were also calibrated offline via~\cite{rehder2016extending}.
Note that although both temporal and spatial parameters between sensors can be estimated online~\cite{li2014online,maye2016online}, reducing these uncertainties beforehand is beneficial for the overall performance.
	


To test the proposed method in different environments, we collected three datasets in indoor environments and three data sets on urban streets.
During the data collection, 
images were recorded at $10$Hz with $640\times400$ and $1280\times800$ pixel resolutions in indoor and outdoor environments respectively.
Measurements from all IMUs were at $200$Hz. For indoor data sets, the trajectory lengths are around $50$ to $100$ meters.
For the urban street data sets, the lengths are about $3.5$km.
For the tests on urban streets, we also recorded the poses from the RTK-GPS, which are used as ground truth for computing RMS errors for different algorithms.
In indoor environments, due to lack of precise ground truth, we started and stopped the motion of our sensor platform at exactly the same location, to enable computing the final drift as the error metric.
Similar to the simulation tests, we here also implemented VIO algorithms by using different number of IMUs as well as the competing algorithm~\cite{eckenhoff2019sensor}.

\subsubsection{Qualitative Results}
We here report the qualitative results of all methods tested in both indoor and urban street tests, which are shown in Fig.~\ref{fig:exp error}.
For all datasets, the proposed method consistently outperforms the original VIO algorithm with a single IMU. In fact, the average error reduction is about $31\%$, which is significant and can be used as guidance when designing hardware sensor platform and software package for performing high-precision localization. 
In addition, when compared against~\cite{eckenhoff2019sensor}, the proposed method yields similar localization accuracy, since both methods utilize measurement information from multiple IMUs without non-probabilistic approximations.

Moreover, the computational efficiency of the proposed method and~\cite{eckenhoff2019sensor} is evaluated with the following metric:
\begin{align}
\label{eq:exp eq}
\kappa \!\!=\!\! \frac{\text{Err. in one-IMU method} - \text{Err. in testing method}}{\text{Time used per update cycle in testing method}}
\end{align}
In fact, $\kappa$ represents the gain in localization accuracy per computational cost, and the results are shown in Table~\ref{table: pose_time}.
The results demonstrate that, the proposed method is significantly more efficient when seeking for accuracy gain. 
In addition, the average time per update for tested algorithms is also given by Table~\ref{table: pose_time raw}, which shows that the computational cost of the proposed method is almost identical to that of the single-IMU method and that of~\cite{eckenhoff2019sensor} is noticeably higher.
This verifies our theoretical analysis that since~\cite{eckenhoff2019sensor} models all IMUs' poses in the state vector of an estimator, the computational cost grows significantly along with the number of IMU. By contrast, in our method, since we generate data for a virtual IMU before performing update (with linear computational cost in the number of IMU, and constant cost in the number of VIO keyframes), the computational cost almost keeps unchanged.

 
It is also interesting to point out that the accuracy gain achieved in our test is similar to that of~\cite{leutenegger2015keyframe}, which uses an 11-cm baseline stereo vision system.
Compared to stereo solution, our method is of lower sensor cost and computational cost. Moreover, the performance of stereo based algorithms will degrade once stereo baseline becomes shorter~\cite{herath2006modeling}. The feasible baseline for various applications, e.g., mobile devices, will be shorter than 11cm, and the proposed method will become even more preferred in those cases. Moreover, compared to using a single IMU with higher precision, the proposed method has comparable localization accuracy yet lower sensor cost.
 \begin{table}
	\centering  
	\begin{threeparttable}  
		\caption{Average accuracy improvement per computational cost, computed by Eq.~\ref{eq:exp eq}.}  \label{table: pose_time}
		\begin{tabular}{l p{1.5cm}<{\centering}
				p{1.3cm}<{\centering}
				p{1.3cm}<{\centering}
			}  
			
			\toprule  
			\quad Num. Seq & 1 & 2 & 3 \\
			
			\midrule
			\quad Eckenhoff et al. \cite{eckenhoff2019sensor}  & 0.2545  & 0.1186 &  0.2865 \\
			\quad Proposed & 0.6894  & 0.3795 &  0.8198 \\
			\bottomrule  
		\end{tabular}  
	\end{threeparttable}  
\end{table} 

 \begin{table}
	\centering  
	\begin{threeparttable}  
		\caption{Time (msec) per update for competing algorithms}  \label{table: pose_time raw}
		\begin{tabular}{l p{1.3cm}<{\centering}
				p{1.1cm}<{\centering}
				p{1.1cm}<{\centering}
			}  
			
			\toprule  
			\quad Num. Seq & 1 & 2 & 3 \\
			
			\midrule
			\quad Single-IMU (msec) & 9.9380 & 9.4280 & 10.6717 \\ 
			\quad Eckenhoff et al. \cite{eckenhoff2019sensor}  (msec) & 39.5155  & 40.9543 &  42.0586 \\
			\quad Proposed (msec) & 10.0381  & 9.8014 &  11.6740 \\
			\bottomrule  
		\end{tabular}  
	\end{threeparttable}  
\end{table} 




\section{CONCLUSIONS} \label{sec:conclusion}
In this paper, we present a lightweight and accurate localization algorithm by using multiple inertial measurement units and exteroceptive sensors. 
Specifically, we propose a method of optimally generating a virtual IMU from the measurements of multiple IMUs, which is followed by closed-form derivation of virtual IMU propagation equations for sensor fusion. By integrating the proposed method into a VIO algorithm, we show that it significantly improves the localization accuracy and outperforms competing algorithms by wide margins.

\appendix
We here provide detailed derivation for 
$\bm \Psi$ and $\bm \Xi$:
\begin{align}
\bm \Psi\!\!=\!\!
\frac{\partial {\bf S} ({\bm \omega}_{mV} \!\!-\!\! {^V{\bm \omega}})}{\partial \bm {b}_{gV}},\,\,
\bm \Xi\!\!=\!\!
\frac{\partial {\bf S} ({\bm \omega}_{mV} \!\!-\!\! {^V{\bm \omega}})}{\partial \bm {n}_{gV}}
\end{align}
To start with, we write $\hat{{\bm \zeta} }$ from Eq.~\ref{eq:zeta} as:
\begin{align}
\hat{\bm \zeta} & \simeq \!-\!{\lfloor\hat{\bm {b}}_{gV} \rfloor}^2
\!\!+\!
{\lfloor\bm {\omega}_{mV}  \rfloor}
{\lfloor\hat{\bm {b}}_{gV} \rfloor}\!+\!{\lfloor\hat{\bm {b}}_{gV} \rfloor}
{\lfloor\bm {\omega}_{mV}  \rfloor}
\label{eq:zeta est}
\end{align}  
Subtracting Eq.~\ref{eq:zeta} from Eq.~\ref{eq:zeta est} leads to
\begin{align}
\tilde{\bm \zeta} &= {\bm \zeta} - \hat{\bm \zeta}  \simeq
\!-\!
{\lfloor \hat{\bm {b}}_{gV} \rfloor}
{\lfloor \tilde{\bm {b}}_{gV} \rfloor}
\!-\!
{\lfloor \tilde{\bm {b}}_{gV} \rfloor}
{\lfloor \hat{\bm {b}}_{gV} \rfloor}
\!\!+\!
{\lfloor\bm {\omega}_{mV}  \rfloor}
{\lfloor\tilde{\bm {b}}_{gV} \rfloor} \notag \\ 
&+\!
{\lfloor\bm {n}_{gV} \rfloor}
{\lfloor\bm {\omega}_{mV}  \rfloor}\!-\!
{\lfloor\hat{\bm {b}}_{gV} \rfloor}
{\lfloor\bm {\ n}_{gV} \rfloor} 
+{\lfloor\tilde{\bm {b}}_{gV} \rfloor}
{\lfloor\bm {\omega}_{mV}  \rfloor}\notag \\ 
&+\!
{\lfloor\bm {\omega}_{mV}  \rfloor}
{\lfloor\bm {n}_{gV} \rfloor}\!-\!
{\lfloor\bm {\ n}_{gV} \rfloor}
{\lfloor\hat{\bm {b}}_{gV} \rfloor}
\notag \\
&= {\lfloor{}^V \hat{\bm {\omega}}  \rfloor}
{\lfloor\tilde{\bm {b}}_{gV} \rfloor} \!\!+\!\!
{\lfloor\tilde{\bm {b}}_{gV} \rfloor}
{\lfloor{}^V \hat{\bm {\omega}}  \rfloor} \!\!+\!\!
 {\lfloor{}^V \hat{\bm {\omega}}  \rfloor} 
{\lfloor\bm {n}_{gV} \rfloor} \!\!+\!\!
{\lfloor\bm {n}_{gV} \rfloor}
{\lfloor{}^V \hat{\bm {\omega}}  \rfloor} \notag 
\end{align}  
Therefore, for any vector $\bm y$, we have
\begin{align}
\tilde{\bm \zeta} \bm y &\!\!= \!\!-\!\!\left(\!
 {\lfloor{}^V \hat{\bm {\omega}}  \rfloor}
{\lfloor {\bm y}\rfloor} \!\!+\!\!
\lfloor 
{\lfloor{}^V \hat{\bm {\omega}}  \rfloor} \bm y \rfloor\! \right)\! \tilde{\bm {b}}_{gV} 
\!\!-\!\!
\left(\!
 {\lfloor{}^V \hat{\bm {\omega}}  \rfloor}
{\lfloor {\bm y}\rfloor} \!\!+\!\!
\lfloor 
{\lfloor{}^V \hat{\bm {\omega}}  \rfloor} \bm y \rfloor \!\right) \!\bm {n}_{gV}.  \notag
\end{align}
As a result,
\begin{align}
{\bf S} ({\bm \omega}_{mV} - {^V{\bm \omega}})
 &= \begin{bmatrix}
{^A}{\Rot}_V {\bm \zeta}  {^V}{{\bm p}}{_A}\\
{^B}{\Rot}_V {\bm \zeta}  {^V}{{\bm p}}{_B}
\end{bmatrix} =
\begin{bmatrix}
{^A}{\Rot}_V ( \hat{\bm \zeta} + \tilde{\bm \zeta}  ) {^V}{{\bm p}}{_A}\\
{^B}{\Rot}_V ( \hat{\bm \zeta} + \tilde{\bm \zeta} )   {^V}{{\bm p}}{_B}
\end{bmatrix}
\notag \\
&= \begin{bmatrix}
{^A}{\Rot}_V \hat{\bm \zeta}  {^V}{{\bm p}}{_A}\\
{^B}{\Rot}_V \hat{\bm \zeta}  {^V}{{\bm p}}{_B}
\end{bmatrix} + {\bm \Psi} \tilde{\bm {b}}_{gV}  + {\bm \Xi} \bm {n}_{gV} 
\end{align}
where 
\begin{align}
{\bm \Psi} = {\bm \Xi} = 
\begin{bmatrix}
{^A}{\Rot}_V \left( -
 {\lfloor{}^V \hat{\bm {\omega}}  \rfloor}
{\lfloor {{^V}{{\bm p}}{_A}}\rfloor} -
\lfloor 
{\lfloor{}^V \hat{\bm {\omega}}  \rfloor} {^V}{{\bm p}}{_A} \rfloor \right)  \\
{^B}{\Rot}_V \left( -
 {\lfloor{}^V \hat{\bm {\omega}}  \rfloor}
{\lfloor {{^V}{{\bm p}}{_B}}\rfloor} -
\lfloor 
{\lfloor{}^V \hat{\bm {\omega}}  \rfloor} {^V}{{\bm p}}{_B} \rfloor \right).
\end{bmatrix} 
\end{align}
This completes our derivation.

\bibliographystyle{IEEEtran}
\bibliography{ref_all.bib}

\end{document}